\documentclass[conference]{IEEEtran}
\IEEEoverridecommandlockouts
\usepackage{graphicx,subfigure,amsmath,amsfonts,amssymb,algorithm,algorithmic,cite,multirow,mathrsfs,booktabs,threeparttable}
\makeatletter

\newcommand{\Rmnum}[1]{\expandafter\@slowromancap\romannumeral #1@}
\makeatother

\def\BibTeX{{\rm B\kern-.05em{\sc i\kern-.025em b}\kern-.08em
    T\kern-.1667em\lower.7ex\hbox{E}\kern-.125emX}}

\begin{document}
\title{Supervised Discriminative Sparse PCA with Adaptive Neighbors for Dimensionality Reduction}

\author{\IEEEauthorblockN{Zhenhua Shi, Dongrui Wu, Jian Huang}
\IEEEauthorblockA{\textit{School of Artificial Intelligence and Automation}\\
\textit{Huazhong University of Science and Technology}\\
Wuhan, China\\
Email: \{zhenhuashi, drwu, huang\_jan\}@hust.edu.cn}\and
\IEEEauthorblockN{Yu-Kai Wang, Chin-Teng Lin}
\IEEEauthorblockA{\textit{Faculty of Engineering and Information Technology}\\
\textit{University of Technology}\\
Sydney, Australia\\
Email: \{YuKai.Wang, Chin-Teng.Lin\}@uts.edu.au}
}

\maketitle

\begin{abstract}
Dimensionality reduction is an important operation in information visualization, feature extraction, clustering, regression, and classification, especially for processing noisy high dimensional data. However, most existing approaches preserve either the global or the local structure of the data, but not both. Approaches that preserve only the global data structure, such as principal component analysis (PCA), are usually sensitive to outliers. Approaches that preserve only the local data structure, such as locality preserving projections, are usually unsupervised (and hence cannot use label information) and uses a fixed similarity graph. We propose a novel linear dimensionality reduction approach, supervised discriminative sparse PCA with adaptive neighbors (SDSPCAAN), to integrate neighborhood-free supervised discriminative sparse PCA and projected clustering with adaptive neighbors. As a result, both global and local data structures, as well as the label information, are used  for better dimensionality reduction. Classification experiments on nine high-dimensional datasets validated the effectiveness and robustness of our proposed SDSPCAAN.
\end{abstract}

\IEEEpeerreviewmaketitle

\section{Introduction}

Defined as the process of projecting high-dimensional data into a low-dimensional subspace, dimensionality reduction is an important operation in information visualization, feature extraction, clustering, regression, and classification \cite{Chao2019}. Linear dimensionality reduction approaches are frequently used for processing noisy high dimensional data, due to their low computational cost and simple geometric interpretations \cite{Cunningham2015}. We divide linear dimensionality reduction approaches into three groups: neighborhood-free approaches, fixed-neighborhood based approaches, and adaptive-neighborhood based approaches.

Neighborhood-free dimensionality reduction approaches require no neighborhood information. For instance, as one of the first dimensionality reduction approaches in the literature, principal component analysis (PCA) \cite{Pearson1901} reduces the dimensionality of data by projecting them onto orthogonal directions of high variances. The traditional PCA is unsupervised, and hence cannot make use of label information. To extend it to supervised learning, supervised PCA \cite{Barshan2011} maximizes the Hilbert-Schmidt independence between the labels and the orthogonally projected data. To incorporate PCA with Laplacian eigenmaps \cite{Belkin2001}, graph-Laplacian PCA (gLPCA) \cite{Jiang2013} adds a weighted Laplacian embedding loss to a variant formulation of PCA (vPCA) for closed-form solution. To extend vPCA to supervised sparse learning, supervised discriminative sparse PCA (SDSPCA) \cite{Feng2019} adds a label-related term and a sparse $L_{2,1}$ regularization \cite{Nie2010} to vPCA. As one of the most widely used supervised dimensionality reduction approaches, linear discriminant analysis (LDA) \cite{Fisher1936} seeks for directions of high separation between different classes. Robust LDA \cite{Zhao2019} reformulates the traditional LDA by minimizing the within class covariance and reducing the influence of outliers via $L_{2,1}$ regularization \cite{Nie2010}. Self-weighted adaptive locality discriminant analysis \cite{Guo2018} reformulates the traditional LDA by minimizing the within class covariance in a pairwise expression and adding $L_{2,1}$ regularization \cite{Nie2010}. To utilize multi-view data in dimensionality reduction, canonical correlations analysis (CCA) \cite{Hotelling1936} jointly maps data from two sources into the same subspace and maximizes the correlation between the projected data. To extend CCA to supervised learning, discriminative CCA \cite{Sun2008} maximizes the within-class correlation between the projected data. Discriminative sparse generalized CCA \cite{drwuCAC2019} further adds sparsity to the discriminative CCA, and also extends it to more than two views.

To preserve neighborhood information in dimensionality reduction, fixed-neighborhood based dimensionality reduction approaches usually assign a fixed similarity graph of data via heat kernel, nearest neighbors, reconstruction weights \cite{He2005}, or local scaling \cite{Zelnik-Manor2004}. For example, as a linear approximation of the Laplacian eigenmaps \cite{Belkin2001}, locality preserving projections (LPP) \cite{He2003} forces the projection of the connected points in the similarity graph to stay as close as possible. As a linear analogy to locally linear embedding \cite{Roweis2000}, neighborhood preserving embedding (NPE) \cite{He2005} represents each data point as a linear combination of its neighbors, and then forces the projection of the points to preserve this reconstruction relationship. Local Fisher discriminant analysis \cite{Sugiyama2007} reformulates LDA in a pairwise expression and assigns a weight to each pairwise distance via a similarity graph. Similarly, locality preserving CCA \cite{Sun2007} reformulates CCA in a pairwise expression and adds weights to the pairwise distances via a similarity graph.

Different from fixed-neighborhood based dimensionality reduction approaches that rely on a fixed similarity graph, adaptive-neighborhood based dimensionality reduction approaches use an adaptive similarity graph. For instance, projected clustering with adaptive neighbors (PCAN) \cite{Nie2014} allows for adaptive neighbors and is able to construct a predefined number of clusters via graphs \cite{Mohar1991,Chung1997}. To extend PCAN to multi-view learning, multi-view feature extraction with structured graph \cite{Zhuge2017} minimizes the differences between the adaptive similarity graph of all views and the fixed similarity graph of each view. To use PCAN for feature selection, structured optimal graph feature selection \cite{Nie2019} adds a weighted $L_{2,p}$ regularization of orthogonal projection matrix to the objective function of PCAN. Projective unsupervised flexible embedding with optimal graph \cite{Wang2018} combines PCAN and ridge regression for image and video representation. To extend PCAN to supervised learning, simultaneously learning neighborship and projection (SLNP) \cite{Pang2019} learns class-wise similarity graphs and the projection matrix simultaneously.

In summary, neighborhood-free dimensionality reduction approaches that preserve the global data structure are usually more sensitive to outliers than neighborhood based approaches that preserve the local structure. To remedy this, SDSPCA \cite{Feng2019} applies $L_{2,1}$ regularization \cite{Nie2010} to reduce the influence of outliers. Adaptive-neighborhood based dimensionality reduction approaches that learn the similarity graph and projection matrix simultaneously are usually advantageous to fixed-neighborhood based approaches. However, most existing adaptive-neighborhood based dimensionality reduction approaches are unsupervised, leading to unsatisfactory classification performance. To remedy this, SLNP \cite{Pang2019} extends PCAN to supervised learning, but it needs adequate data from each class for class-wise similarity graph construction.

This paper proposes supervised discriminative sparse PCA with adaptive neighbors (SDSPCAAN) for dimensionality reduction, which extends PCAN to supervised learning, following the approach in \cite{Nie2017}, and integrates it with the state-of-the-art SDSPCA.
%Its main contributions are:
%\begin{enumerate}
%\item We extend PCAN to supervised learning so that it can preserve more discriminative information in dimensionality reduction.
%
%\item We propose SDSPCAAN that unifies PCAN and SDSPCA to utilize both global and local data structure information.
%
%\item We conduct extensive classification experiments on nine high-dimensional datasets to validate the effectiveness and robustness of our proposed SDSPCAAN.
%\end{enumerate}

The remainder of this paper is organized as follows: Section~II introduces PCA, SDSPCA, PCAN and our proposed SDSPCAAN. Section~III describes the nine high-dimensional datasets and our experimental results. Section~IV draws conclusions.

\section{Methods}

In this paper, matrices and vectors are denoted by uppercase and lowercase boldface letters, respectively. Other important notations are summarized in Table~\ref{tab:notations}.

\begin{table}[htpb]\centering \setlength{\tabcolsep}{1mm}
\caption{Notations used in this paper.} \label{tab:notations}
\begin{tabular}{@{}l|l@{}}
\toprule
Notation                                & \multicolumn{1}{c}{Meaning}                                                               \\ \midrule
$\mathbf{S}\in \mathbb{R}^{n \times n}$ & The pairwise similarity matrix of $\mathbf{X}$                                            \\
%$\mathbf{s}_i$                          & The $i$th row with transpose of matrix $\mathbf{S}$     \\
$S_{ij}$                                & The $(i,j)$th element of matrix $\mathbf{S}$                                              \\
$\mathbf{S}^T$                          & The transpose of matrix $\mathbf{S}$                                                      \\
$\operatorname{Tr}(\mathbf{S})$         & The trace of a square matrix $\mathbf{S}$                                                 \\
$\|\mathbf{Q}\|_F$                      & The Frobenius norm of matrix $\mathbf{Q}$                                                 \\
$\|\mathbf{Q}\|_{1,1}$                  & The $L_{1,1}$ norm of matrix $\mathbf{Q}\in \mathbb{R}^{n \times k}$,                     \\
                                        & i.e., $\|\mathbf{Q}\|_{1,1}=\sum_{i=1}^n\left(\sum_{j=1}^k \|Q_{ij}\|\right)$             \\
$\|\mathbf{Q}\|_{2,1}$                  & The $L_{2,1}$ norm of matrix $\mathbf{Q}\in \mathbb{R}^{n \times k}$,                     \\
                                        & i.e., $\|\mathbf{Q}\|_{2,1}=\sum_{i=1}^n\left(\sum_{j=1}^k Q_{ij}^2\right)^{\frac{1}{2}}$ \\
$\|\mathbf{s}_i\|_2$                    & The $L_{2}$ norm of vector $\mathbf{s}_i$                                                 \\
$\operatorname{diag}(\mathbf{v})$       & The square diagonal matrix with the elements                                              \\
                                        & of vector $\mathbf{v}$ on the main diagonal                                               \\
$\mathbf{I}_k$                          & A $k \times k$ identity matrix                                                            \\
$\mathbf{I}_{d\times k}$                & A $d \times k$ matrix with ones in the main diagonal                                      \\
                                        & and zeros elsewhere                                                                       \\
$\mathbf{1}_{n\times k}$                & A $n \times k$ all-one matrix                                                             \\
$\mathbf{0}_{n\times k}$                & A $n \times k$ all-zero matrix                                                            \\
$\mathbf{W}\in \mathbb{R}^{d \times k}$ & The subspace projection matrix of $\mathbf{X}$,                                           \\
                                        & where $k$ is the subspace dimensionality                                                  \\
$\mathbf{Q}\in \mathbb{R}^{n \times k}$ & The auxiliary matrix of $\mathbf{W}$                                                      \\
$\mathbf{L}\in \mathbb{R}^{n \times n}$ & The Laplacian matrix of $\mathbf{S}\in \mathbb{R}^{n \times n}$,                                                     \\
                                        & i.e., $\mathbf{L}=\operatorname{diag}(\mathbf{S}\mathbf{1}_{n\times 1})-\mathbf{S}$       \\ \bottomrule
\end{tabular}
\end{table}

The training data matrix is $\mathbf{X}=\left[\mathbf{x}_{1}, \ldots, \mathbf{x}_{n}\right]^T \in \mathbb{R}^{n \times d}$, where $n$ is the number of training samples, and $d$ the feature dimensionality. Without loss of generality, we assume $\mathbf{X}$ is mean-centered, i.e., $\mathbf{1}_{1\times n}\mathbf{X}=\mathbf{0}_{1\times d}$. The one-hot coding label matrix of $\mathbf{X}$ is $\mathbf{Y}=\left[\mathbf{y}_{1}, \ldots, \mathbf{y}_{n}\right]^T\in \mathbb{R}^{n \times c}$, where $c$ is the number of classes.

\subsection{Principal Component Analysis (PCA)} \label{sect:PCA}

PCA \cite{Pearson1901} reduces the dimensionality of data by projecting them onto orthogonal directions of high variances, which usually have higher signal-to-noise ratios than the directions of low variances \cite{DeBie2005}. Mathematically, it solves the following optimization problem:
\begin{align}
\max_{\mathbf{W}} \operatorname{Tr}(\mathbf{W}^T\mathbf{X}^T\mathbf{X}\mathbf{W}) \quad
\rm{s.t.}\ \mathbf{W}^T\mathbf{W}=\mathbf{I}_{k}. \label{eq:MaxVar}
\end{align}
The optimal $\mathbf{W}\in \mathbb{R}^{d \times k}$ is formed by the $k$ leading eigenvectors of $\mathbf{X}^T\mathbf{X}$, which also minimizes the Frobenius norm of the residual matrix:
\begin{align}
\min_{\mathbf{W}} \|\mathbf{X}-\mathbf{X}\mathbf{W}\mathbf{W}^T\|_F^2 \quad
\rm{s.t.}\ \mathbf{W}^T\mathbf{W}=\mathbf{I}_{k}. \label{eq:MinRes}
\end{align}

A variant formulation of PCA (vPCA) used in gLPCA \cite{Jiang2013} and SDSPCA \cite{Feng2019} optimizes
\begin{align}
\begin{split}
&\min_{\mathbf{Q}} \|\mathbf{X}-\mathbf{Q}\mathbf{Q}^T\mathbf{X}\|_F^2
=\max_{\mathbf{Q}}\operatorname{Tr}(\mathbf{Q}^T\mathbf{X}\mathbf{X}^T\mathbf{Q}) \\
&\rm{s.t.}\ \mathbf{Q}^T\mathbf{Q}=\mathbf{I}_{k}. \label{eq:PCA2}
\end{split}
\end{align}
The optimal $\mathbf{Q}\in \mathbb{R}^{n \times k}$ is formed by the $k$ leading eigenvectors of $\mathbf{X}\mathbf{X}^T$. The projection matrix is then $\mathbf{W}=\mathbf{X}^T\mathbf{Q}$.

Let $\mathbf{X}=\mathbf{U}\mathbf{\Sigma}\mathbf{R}^T$ be the singular value decomposition of $\mathbf{X}$, where $\mathbf{U}\in \mathbb{R}^{n \times n}$ and $\mathbf{R}\in \mathbb{R}^{d \times d}$ are orthogonal, and $\mathbf{\Sigma}\in \mathbb{R}^{n \times d}$ is a diagonal matrix with non-negative singular values in descending order on the diagonal. Then, we can calculate the optimal projection matrix for PCA as $\mathbf{W}_{\text{PCA}}=\mathbf{R}_{1:k}$, the optimal projection matrix for vPCA as $\mathbf{W}_{\text{vPCA}}=\mathbf{R}_{1:k}\mathbf{\Sigma}_{1:k}$, where $\mathbf{R}_{1:k}\in \mathbb{R}^{d \times k}$ consists of the first $k$ columns of $\mathbf{R}$, and $\mathbf{\Sigma}_{1:k}\in \mathbb{R}^{k \times k}$ is a diagonal matrix of the first $k$ leading singular values (arranged in descending order). Thus,
\begin{align}\label{eq:EqualPCA}
\mathbf{W}_{\text{vPCA}}=\mathbf{W}_{\text{PCA}}\mathbf{\Sigma}_{1:k}.
\end{align}

vPCA is equivalent to PCA if we scale each column of $\mathbf{XW}$ by the column standard deviation, which is a common practice in machine learning.

\subsection{Supervised Discriminative Sparse PCA (SDSPCA)} \label{sect:SDSPCA}

SDSPCA \cite{Feng2019} extends vPCA to supervised sparse linear dimensionality reduction, by integrating data information, label information and sparse regularization. The projection matrix $\mathbf{W}$ is obtained by
\begin{align}\label{eq:SDSPCA}
\begin{split}
\min_{\mathbf{W},\mathbf{G},\mathbf{Q}} &\|\mathbf{X}-\mathbf{Q}\mathbf{W}^T\|_F^2+\alpha \|\mathbf{Y}-\mathbf{Q}\mathbf{G}^T\|_F^2+\beta \|\mathbf{Q}\|_{2,1}  \\
\rm{s.t.}\ & \mathbf{Q}^T\mathbf{Q}=\mathbf{I}_{k},
\end{split}
\end{align}
where $\mathbf{G}\in \mathbb{R}^{c \times k}$, and $\alpha$ and $\beta$ are scaling weights. Alternating optimization can be used to solve (\ref{eq:SDSPCA}), as follows.

When $\mathbf{G}$ and $\mathbf{Q}$ are fixed, setting the partial derivative of (\ref{eq:SDSPCA}) w.r.t. $\mathbf{W}$ to zero yields
\begin{align}
\mathbf{W} = \mathbf{X}^T\mathbf{Q}. \label{eq:SDSPCA-W}
\end{align}

When $\mathbf{Q}$ and $\mathbf{W}$ are fixed, similarly, we have
\begin{align}
\mathbf{G} = \mathbf{Y}^T\mathbf{Q}. \label{eq:SDSPCA-G}
\end{align}

When $\mathbf{W}$ and $\mathbf{G}$ are fixed, substituting (\ref{eq:SDSPCA-W}) and (\ref{eq:SDSPCA-G}) into (\ref{eq:SDSPCA}) yields
\begin{align}\label{eq:SDSPCA-Q}
\begin{split}
\min_{\mathbf{Q}} &\|\mathbf{X}-\mathbf{Q}\mathbf{Q}^T\mathbf{X}\|_F^2+\alpha \|\mathbf{Y}-\mathbf{Q}\mathbf{Q}^T\mathbf{Y}\|_F^2+\beta \|\mathbf{Q}\|_{2,1}  \\
=\min_{\mathbf{Q}} &-\operatorname{Tr}(\mathbf{Q}^T\mathbf{X}\mathbf{X}^T\mathbf{Q})
-\alpha\operatorname{Tr}(\mathbf{Q}^T\mathbf{Y}\mathbf{Y}^T\mathbf{Q})
+\beta\operatorname{Tr}(\mathbf{Q}^T\mathbf{D}\mathbf{Q})\\
=\min_{\mathbf{Q}} &\operatorname{Tr}\left(\mathbf{Q}^T(-\mathbf{X}\mathbf{X}^T-\alpha\mathbf{Y}\mathbf{Y}^T+\beta\mathbf{D})\mathbf{Q}\right) \\
\rm{s.t.}\ & \mathbf{Q}^T\mathbf{Q}=\mathbf{I}_{k}.
\end{split}
\end{align}
The optimal $\mathbf{Q}$ is formed by the $k$ trailing eigenvectors of $\mathbf{Z}=-\mathbf{X}\mathbf{X}^T-\alpha\mathbf{Y}\mathbf{Y}^T+\beta\mathbf{D}$, where $\mathbf{D}\in \mathbb{R}^{n \times n}$ is a diagonal matrix with the $i$th diagonal element be \cite{Nie2010}
\begin{align}
D_{ii} = \frac{1}{2\sqrt{\sum_{j=1}^k Q_{ij}^2+\epsilon}}, \label{eq:SDSPCA-V}
\end{align}
where $\epsilon$ is a small positive constant to avoid dividing by zero.

The pseudocode for optimizing SDSPCA is shown in Algorithm~\ref{Alg:SDSPCA}.

\begin{algorithm}
\caption{The SDSPCA training algorithm \cite{Feng2019}.}\label{Alg:SDSPCA}
\begin{algorithmic}
\REQUIRE $\mathbf{X}\in \mathbb{R}^{n\times d}$, the training data matrix; \\
\hspace*{8mm} $\mathbf{Y}\in \mathbb{R}^{n\times c}$, the one-hot coding label matrix of $\mathbf{X}$;\\
\hspace*{8mm} $k$, the subspace dimensionality; \\
\hspace*{8mm} $\alpha$ and $\beta$, the scaling weights; \\
\hspace*{8mm} $\epsilon$, a small positive constant; \\
\hspace*{8mm} $tol$, the tolerance; \\
\hspace*{8mm} $T$, the maximum number of iterations.
\ENSURE Projection matrix $\mathbf{W}\in \mathbb{R}^{d \times k}$.
\STATE Initialize $\mathbf{Z}_0=-\mathbf{X}\mathbf{X}^T-\alpha\mathbf{Y}\mathbf{Y}^T$, $\mathbf{D}=\mathbf{I}_{n}$, and $\mathbf{Q}_0=\mathbf{0}_{n\times k}$;
\FOR{$t=1$ to $T$}
\STATE Calculate $\mathbf{Z}=\mathbf{Z}_0+\beta\mathbf{D}$;
\STATE Construct $\mathbf{Q}$ by the $k$ trailing eigenvectors of $\mathbf{Z}$;
\IF{$\|\mathbf{Q}-\mathbf{Q}_0\|_{1,1} < tol$}
\STATE break;
\ENDIF
\STATE Update $\mathbf{D}$ using (\ref{eq:SDSPCA-V});
\STATE $\mathbf{Q}_0=\mathbf{Q}$;
\ENDFOR
\STATE Calculate $\mathbf{W}$ using (\ref{eq:SDSPCA-W}).
\end{algorithmic}
\end{algorithm}

SDSPCA integrates data information and label information elegantly to seek for a discriminative low-dimensional subspace, and it does not involve any matrix inversion operation. Additionally, the sparse constraint of $\mathbf{Q}$ makes it robust to outliers. However, SDSPCA fails to utilize the neighborhood information.

\subsection{Projected Clustering with Adaptive Neighbors (PCAN)}

Different from SDSPCA, which ignores neighborhood information, PCAN \cite{Nie2014} learns the projection matrix and neighbourhood relations simultaneously to perform unsupervised linear dimensionality reduction. Its projection matrix $\mathbf{W}$ is obtained by:
\begin{align}\label{eq:PCAN}
\begin{split}
&\min _{\mathbf{W}, \mathbf{F}, \mathbf{S}} \sum_{i, j=1}^{n}\left(\|\mathbf{W}^T\mathbf{x}_{i}-\mathbf{W}^T\mathbf{x}_{j}\|_{2}^{2} S_{i j}+\gamma_i S_{ij}^2+\lambda\|\mathbf{f}_i-\mathbf{f}_j\|_2^2S_{ij} \right)\\
&=\min _{\mathbf{W}, \mathbf{F}, \mathbf{S}} 2\operatorname{Tr}\left(\mathbf{W}^T \mathbf{X}^T\mathbf{L} \mathbf{X}\mathbf{W}\right)+\operatorname{Tr}(\mathbf{S}^T\mathbf{\Gamma}\mathbf{S})+2\lambda \operatorname{Tr}\left(\mathbf{F}^{T} \mathbf{L} \mathbf{F}\right)\\
&\rm{s.t.}\ \mathbf{S} \mathbf{1}_{n\times1}=\mathbf{1}_{n\times1}, \mathbf{S} \geq 0, \mathbf{W}^T\mathbf{X}^T\mathbf{X}\mathbf{W}=\mathbf{I}_{k},\mathbf{F}^{T} \mathbf{F}=\mathbf{I}_c,
\end{split}
\end{align}
where $\mathbf{S}\in \mathbb{R}^{n \times n}$ is the pairwise similarity matrix, $\mathbf{L}=\operatorname{diag}(\mathbf{S1}_{n\times1})-\mathbf{S}\in \mathbb{R}^{n \times n}$ is the Laplacian matrix, $\mathbf{\Gamma}\in \mathbb{R}^{n \times n}$ is a diagonal matrix with the $i$th diagonal element being $\gamma_i$, $\lambda$ is a scaling weight, $\mathbf{F} \in \mathbb{R}^{n \times c}$ is an auxiliary matrix for minimizing the $c$ smallest eigenvalues of $\mathbf{L}$. Since the multiplicity of $0$ as an eigenvalue of $\mathbf{L}$ is equal to the number of connected components of $\mathbf{S}$ \cite{Mohar1991,Chung1997}, a proper $\lambda$ will lead to exactly $c$ clusters indicated by $\mathbf{S}$. Alternating optimization can be used to solve (\ref{eq:PCAN}), as follows.

When $\mathbf{F}$ and $\mathbf{S}$ are fixed, (\ref{eq:PCAN}) becomes
\begin{align}\label{eq:PCAN-W}
\min _{\mathbf{W}} \operatorname{Tr}\left(\mathbf{W}^T \mathbf{X}^T\mathbf{L} \mathbf{X}\mathbf{W}\right) \quad
\rm{s.t.}\ \mathbf{W}^T\mathbf{X}^T\mathbf{X}\mathbf{W}=\mathbf{I}_k.
\end{align}
The optimal $\mathbf{W}$ is formed by the $k$  trailing eigenvectors of $(\mathbf{X}^T\mathbf{X})^{-1}\mathbf{X}^T\mathbf{L} \mathbf{X}$.

When $\mathbf{S}$ and $\mathbf{W}$ are fixed, (\ref{eq:PCAN}) becomes
\begin{align}\label{eq:PCAN-F}
\min _{\mathbf{F}} \operatorname{Tr}\left(\mathbf{F}^{T} \mathbf{L} \mathbf{F}\right) \quad
\rm{s.t.}\ \mathbf{F}^{T} \mathbf{F}=\mathbf{I}_k.
\end{align}
The optimal $\mathbf{F}$ is formed by the $c$  trailing eigenvectors of $\mathbf{L}$.

When $\mathbf{W}$ and $\mathbf{F}$ are fixed, (\ref{eq:PCAN}) becomes
\begin{align}\label{eq:PCAN-S}
\begin{split}
\min _{\mathbf{S}} &\sum_{i,j=1}^{n}\left(\|\mathbf{W}^T\mathbf{x}_{i}-\mathbf{W}^T\mathbf{x}_{j}\|_{2}^{2} S_{i j}+\gamma_i S_{ij}^2+\lambda\|\mathbf{f}_i-\mathbf{f}_j\|_2^2S_{ij} \right) \\
\rm{s.t.}\ & \mathbf{S} \mathbf{1}_{n\times 1}=\mathbf{1}_{n\times 1}, \mathbf{S} \geq 0.
\end{split}
\end{align}

Let $d_{ij}^x=\|\mathbf{W}^T\mathbf{x}_{i}-\mathbf{W}^T\mathbf{x}_{j}\|_{2}^{2}$, $d_{ij}^f=\|\mathbf{f}_i-\mathbf{f}_j\|_{2}^{2}$, $\mathbf{d}_{i}\in \mathbb{R}^{n\times 1}$ be a vector with the $j$-th element being $d_{ij}=d_{ij}^x+\lambda d_{ij}^f$, and $\mathbf{s}_i$ be the transpose of the $i$-th row of $\mathbf{S}$. Then, (\ref{eq:PCAN-S}) can be written in a vector form as
\begin{align}\label{eq:PCAN-si}
\min_{\mathbf{s}_i} \gamma_i \|\mathbf{s}_i\|_2^2 + \mathbf{d}_i^T\mathbf{s}_i \qquad
\rm{s.t.}\ \mathbf{s}_{i}^T \mathbf{1}_{n\times 1}=1, \mathbf{s}_{i} \geq 0.
\end{align}
Differentiating the Lagrangian $\mathcal{L}(\mathbf{s}_i,\eta, \mathbf{b}) = \gamma_i \|\mathbf{s}_i\|_2^2 + \mathbf{d}_i^T\mathbf{s}_i-\eta (\mathbf{s}_{i}^{T} \mathbf{1}_{n\times 1}-1)-\mathbf{b}^T\mathbf{s}_{i}$ corresponding to (\ref{eq:PCAN-si}) with respect to $\mathbf{s}_i$ and setting it to zero leads to
\begin{align}
\begin{split}
\mathbf{s}_i = \frac{1}{2\gamma_i}(-\mathbf{d}_i+\eta\mathbf{1}_{n\times 1}+\mathbf{b}),
\end{split}
\end{align}
where $\eta,\mathbf{b} \geq 0$ are the Lagrangian multipliers. According to the Karush-Kuhn-Tucker (KKT) complementary condition, we have
\begin{align}
\mathbf{b}^T\mathbf{s}_{i}=0.
\end{align}
Then we can express the optimal $\mathbf{s}_i$ as
\begin{align} \label{eq:PCAN-si-opt-gamma}
\mathbf{s}_i = \frac{1}{2\gamma_i}(-\mathbf{d}_i+\eta\mathbf{1}_{n\times 1})_{+},
\end{align}
where $(x)_{+}=\max \{0, x\}$.

Without loss of generality, suppose $d_{i1},\ldots,d_{in}$ are ordered in ascending order. If the optimal $\mathbf{s}_i$ has only $m$ nonzero elements, then according to (\ref{eq:PCAN-si}) and(\ref{eq:PCAN-si-opt-gamma}), we have
\begin{align}
\left\{\begin{aligned}
\sum_{j=1}^m \frac{1}{2\gamma_i}(-d_{ij}+\eta)&=1, \\
\frac{1}{2\gamma_i}(-d_{im}+\eta) &>0, \\
\frac{1}{2\gamma_i}(-d_{i,m+1}+\eta) & \leq 0,
\end{aligned}\right.
\end{align}
which lead to
\begin{align} \label{eq:PCAN-eta-gamma}
\eta = \frac{1}{m}(2\gamma_i +\sum_{j=1}^m d_{ij}),
\end{align}
and
\begin{align} \label{eq:PCAN-gammaBound}
\frac{m}{2} d_{im}-\frac12 \sum_{j=1}^m d_{ij} < \gamma_i \leq \frac{m}{2} d_{i,m+1}-\frac12 \sum_{j=1}^m d_{ij}.
\end{align}
Substituting (\ref{eq:PCAN-si-opt-gamma}) and (\ref{eq:PCAN-eta-gamma}) into the objective function in (\ref{eq:PCAN-si}) yields
\begin{align}
\begin{split} \label{eq:PCAN-gammaFunc}
&\gamma_i \|\mathbf{s}_i\|_2^2 + \mathbf{d}_i^T\mathbf{s}_i \\
=&\gamma_i \sum_{j=1}^m \left(\frac{1}{2\gamma_i}(-d_{ij}+\eta)\right)^2 + \sum_{j=1}^m\frac{d_{ij}}{2\gamma_i}(-d_{ij}+\eta) \\
=&\frac{\gamma_i}{m}+\frac{1}{4\gamma_i m}\left((\sum_{j=1}^m d_{ij})^2-m\sum_{j=1}^m d_{ij}^2\right)+\frac1m \sum_{j=1}^m d_{ij},
\end{split}
\end{align}
where $(\sum_{j=1}^m d_{ij})^2 \leq m\sum_{j=1}^m d_{ij}^2$ according to the Cauchy-Buniakowsky-Schwarz inequality. So, the objective function increases monotonously with respect to $\gamma_i$.

Taking $\gamma_{i}$ as a dual variable, according to (\ref{eq:PCAN-gammaBound}) and(\ref{eq:PCAN-gammaFunc}), the optimal $\gamma_i$ can be expressed as
\begin{align}\label{eq:PCAN-gamma-opt}
\gamma_i=\frac{m}{2} d_{i,m+1}-\frac12 \sum_{j=1}^m d_{ij}.
\end{align}

Substituting (\ref{eq:PCAN-eta-gamma}) and (\ref{eq:PCAN-gamma-opt}) into (\ref{eq:PCAN-si-opt-gamma}) yields the optimal $\mathbf{s}_{i}$, which can be expressed as
\begin{align}\label{eq:PCAN-si-opt}
\begin{split}
\mathbf{s}_{i} = \left(\frac{d_{i,m+1}-\mathbf{d}_{i}}{md_{i,m+1}-\sum_{j=1}^m d_{ij}+\epsilon}\right)_{+},
\end{split}
\end{align}
where $\epsilon$ is a small positive constant to avoid dividing by zero.

The detailed optimization routine of PCAN is shown in Algorithm~\ref{Alg:PCAN}.

\begin{algorithm}
\caption{The PCAN training algorithm \cite{Nie2014}.}\label{Alg:PCAN}
\begin{algorithmic}
\REQUIRE $\mathbf{X}\in \mathbb{R}^{n\times d}$, the training data matrix; \\
\hspace*{8mm} $k$, subspace dimensionality; \\
\hspace*{8mm} $c$, number of clusters; \\
\hspace*{8mm} $m$, number of nearest neighbors; \\
\hspace*{8mm} $\epsilon$, a small positive constant; \\
\hspace*{8mm} $tol$, absolute tolerance; \\
\hspace*{8mm} $T$, maximum number of iterations.
\ENSURE Projection matrix $\mathbf{W}\in \mathbb{R}^{d \times k}$.
\STATE Initialize $d_{ij}$ as $\|\mathbf{x}_{i}-\mathbf{x}_{j}\|_{2}^{2}$, and $\mathbf{S}$ using (\ref{eq:PCAN-si-opt});
\STATE $\lambda=1$;
\FOR{$t=1$ to $T$}
\STATE $\mathbf{S}=(\mathbf{S}+\mathbf{S}^T)/2$;
\STATE $\mathbf{L}=\operatorname{diag}(\mathbf{S1}_n)-\mathbf{S}$;
\STATE Construct $\mathbf{W}$ by the $k$ trailing eigenvectors of $(\mathbf{X}^T\mathbf{X})^{-1}\mathbf{X}^T\mathbf{L} \mathbf{X}$ ;
\STATE Construct $\mathbf{F}$ by the $c$ trailing eigenvectors of $\mathbf{L}$;
\STATE Calculate the $c+1$ smallest eigenvalues of $\mathbf{L}$ in ascending order as $e_1, e_2, \ldots ,e_{c+1}$;
\IF{$\sum_{i=1}^c e_i > tol$}
\STATE $\lambda=2\lambda$;
\ELSIF{$\sum_{i=1}^{c+1} e_i < tol$}
\STATE $\lambda=\lambda/2$;
\ELSE
\STATE break;
\ENDIF
\STATE Calculate $d_{ij}=\|\mathbf{W}^T\mathbf{x}_{i}-\mathbf{W}^T\mathbf{x}_{j}\|_{2}^{2}+\lambda\|\mathbf{f}_i-\mathbf{f}_j\|_{2}^{2}$;
\STATE Update $\mathbf{S}$ using (\ref{eq:PCAN-si-opt});
\ENDFOR
\end{algorithmic}
\end{algorithm}

After being updated by (\ref{eq:PCAN-si-opt}), $\mathbf{S}$ is replaced by $(\mathbf{S}+\mathbf{S}^T)/2$ for symmetry. The sum of the $c$ smallest eigenvalues of $\mathbf{L}$, $\sum_{i=1}^c e_i$, is used to restrict the rank of $\mathbf{L}$ since the eigenvalues of the Laplacian matrix $\mathbf{L}$ are non-negative. When $\sum_{i=1}^c e_i > tol$, the rank of $\mathbf{L}$ is larger than $n-c$, and the number of connected components of $\mathbf{S}$ is smaller than $c$ \cite{Mohar1991,Chung1997}, so $\lambda$ is multiplied by 2 to strengthen the impact of $\operatorname{Tr}\left(\mathbf{F}^{T} \mathbf{L} \mathbf{F}\right)$. When $\sum_{i=1}^{c+1} e_i < tol$, the opposite is performed. We did not use a global $\gamma$, whose value is the average of all $\gamma_i$, as in \cite{Nie2014}; instead, we used the optimal $\gamma_i$ to update the $i$-th row of $\mathbf{S}$ ($\mathbf{s}_i^T$) for faster convergence.

PCAN simultaneously learns the projection matrix and neighbourhood relations to perform dimensionality reduction and construct exactly $c$ clusters based on $\mathbf{S}$. However, $\mathbf{X}^T\mathbf{X}$ can be singular, especially for high-dimensional data, so the construction of $\mathbf{W}$ may not be accurate. In addition, PCAN fails to utilize the label information for better discrimination.

\subsection{Supervised Discriminative Sparse PCA with Adaptive Neighbors (SDSPCAAN)} \label{sect:SDSPCAAN}

To take the advantages of SDSPCA and PCAN and avoid their limitations, we propose SDSPCAAN to integrate SDSPCA and PCAN together. Its projection matrix is obtained by
\begin{align}\label{eq:SDSPCAAN}
\begin{split}
&\min_{\mathbf{Q}, \mathbf{S}} \|\mathbf{X}-\mathbf{Q}\mathbf{Q}^T\mathbf{X}\|_F^2+\alpha \|\mathbf{Y}-\mathbf{Q}\mathbf{Q}^T\mathbf{Y}\|_F^2+\beta \|\mathbf{Q}\|_{2,1}\\
&+\frac{1}{2}\delta\left[2\operatorname{Tr}(\mathbf{Q}^T \mathbf{X}\mathbf{X}^T\mathbf{L} \mathbf{X}\mathbf{X}^T\mathbf{Q})+\operatorname{Tr}(\mathbf{S}^T\mathbf{\Gamma}\mathbf{S})+2\lambda \operatorname{Tr}(\mathbf{Y}^{T} \mathbf{L} \mathbf{Y})\right] \\
&\rm{s.t.}\ \mathbf{Q}^T\mathbf{Q}=\mathbf{I}_k, \mathbf{S} \mathbf{1}_{n\times1}=\mathbf{1}_{n\times1}, \mathbf{S} \geq 0,
\end{split}
\end{align}
where $\delta >0$ is a scaling weight.

We construct SDSPCA based on (\ref{eq:SDSPCA-Q}), and PCAN based on (\ref{eq:PCAN}). We replace $\mathbf{W}$ in PCAN with $\mathbf{X}^T\mathbf{Q}$ based on (\ref{eq:SDSPCA-W}) to avoid matrix inversion error, and $\mathbf{F}$ in PCAN with $\mathbf{Y}$ to utilize label information, following \cite{Nie2017}. Alternating optimization can be used to solve (\ref{eq:SDSPCAAN}), as follows.

When $\mathbf{S}$ is fixed, (\ref{eq:SDSPCAAN}) becomes
\begin{align}\label{eq:SDSPCAAN-Q}
\begin{split}
\min_{\mathbf{Q}} &\|\mathbf{X}-\mathbf{Q}\mathbf{Q}^T\mathbf{X}\|_F^2+\alpha \|\mathbf{Y}-\mathbf{Q}\mathbf{Q}^T\mathbf{Y}\|_F^2+\beta \|\mathbf{Q}\|_{2,1}\\
&+\delta\operatorname{Tr}\left(\mathbf{Q}^T \mathbf{X}\mathbf{X}^T\mathbf{L} \mathbf{X}\mathbf{X}^T\mathbf{Q}\right)   \\
=\min_{\mathbf{Q}} &-\operatorname{Tr}(\mathbf{Q}^T\mathbf{X}\mathbf{X}^T\mathbf{Q})
-\alpha\operatorname{Tr}(\mathbf{Q}^T\mathbf{Y}\mathbf{Y}^T\mathbf{Q})
+\beta\operatorname{Tr}(\mathbf{Q}^T\mathbf{D}\mathbf{Q})\\
&+\delta\operatorname{Tr}\left(\mathbf{Q}^T \mathbf{X}\mathbf{X}^T\mathbf{L} \mathbf{X}\mathbf{X}^T\mathbf{Q}\right)   \\
=\min_{\mathbf{Q}} &\operatorname{Tr}\left(\mathbf{Q}^T(-\mathbf{X}\mathbf{X}^T-\alpha\mathbf{Y}\mathbf{Y}^T+\beta\mathbf{D}+\delta \mathbf{X}\mathbf{X}^T\mathbf{L} \mathbf{X}\mathbf{X}^T)\mathbf{Q}\right) \\
\rm{s.t.}\ & \mathbf{Q}^T\mathbf{Q}=\mathbf{I}_k.
\end{split}
\end{align}
The optimal $\mathbf{Q}$ is formed by the $k$ trailing eigenvectors of $\mathbf{Z}=-\mathbf{X}\mathbf{X}^T-\alpha\mathbf{Y}\mathbf{Y}^T+\beta\mathbf{D}+\delta \mathbf{X}\mathbf{X}^T\mathbf{L} \mathbf{X}\mathbf{X}^T$, where $\mathbf{D}\in \mathbb{R}^{n \times n}$ is a diagonal matrix expressed in (\ref{eq:SDSPCA-V}).

When $\mathbf{Q}$ is fixed, (\ref{eq:SDSPCAAN}) becomes
\begin{align}\label{eq:SDSPCAAN-S}
\begin{split}
\min _{\mathbf{S}} &2\operatorname{Tr}\left(\mathbf{Q}^T\mathbf{X}\mathbf{X}^T\mathbf{L} \mathbf{X}\mathbf{X}^T\mathbf{Q}\right) +\operatorname{Tr}(\mathbf{S}^T\mathbf{\Gamma}\mathbf{S})+2\lambda \operatorname{Tr}\left(\mathbf{Y}^{T} \mathbf{L} \mathbf{Y}\right)\\
\rm{s.t.}\ &\mathbf{S} \mathbf{1}_{n\times1}=\mathbf{1}_{n\times1}, \mathbf{S} \geq 0.
\end{split}
\end{align}
Same as in PCAN, the optimal $\mathbf{s}_i$ can be expressed as
\begin{align}\label{eq:SDSPCAAN-si}
\begin{split}
\mathbf{s}_{i} = (\frac{d_{i,m+1}-\mathbf{d}_{i}}{md_{i,m+1}-\sum_{j=1}^m d_{ij}+\epsilon})_{+},
\end{split}
\end{align}
where $d_{ij}=\|\mathbf{Q}^T\mathbf{X}\mathbf{x}_{i}
-\mathbf{Q}^T\mathbf{X}\mathbf{x}_{j}\|_{2}^{2}+\lambda\|\mathbf{y}_i-\mathbf{y}_j\|_{2}^{2}$, and $\epsilon$ is a small positive constant to avoid dividing by zero.

The detailed optimization routine of SDSPCAAN is given in Algorithm~\ref{Alg:SDSPCAAN}. When $\delta$ in SDSPCAAN is set to zero, it degrades to SDSPCA (Section~\ref{sect:SDSPCA}). When $\delta$ in SDSPCAAN is set to infinity, SDSPCAAN degrades to supervised PCAN (SPCAN). When fixing the similarity graph $\mathbf{S}$ at its initial value, SDSPCAAN degrades to SDSPCA-LPP, a combination of SDSPCA and LPP.

\begin{algorithm}
\caption{The proposed SDSPCAAN training algorithm.}\label{Alg:SDSPCAAN}
\begin{algorithmic}
\REQUIRE $\mathbf{X}\in \mathbb{R}^{n\times d}$, the training data matrix; \\
\hspace*{8mm} $\mathbf{Y}\in \mathbb{R}^{n\times c}$, the corresponding one-hot coding label matrix of $\mathbf{X}$;\\
\hspace*{8mm} $k$, subspace dimensionality; \\
\hspace*{8mm} $m$, number of nearest neighbors; \\
\hspace*{8mm} $\alpha$, $\beta$ and $\delta$, scaling weights; \\
\hspace*{8mm} $\epsilon$, small positive constant; \\
\hspace*{8mm} $tol$, absolute tolerance; \\
\hspace*{8mm} $T$, maximum number of iterations.
\ENSURE Projection matrix $\mathbf{W}\in \mathbb{R}^{d \times k}$.
\STATE $\mathbf{Z}_0=-\mathbf{X}\mathbf{X}^T-\alpha\mathbf{Y}\mathbf{Y}^T$;
\STATE $\mathbf{D}=\mathbf{I}_{n}$;
\STATE $\mathbf{Q}_0=\mathbf{0}_{n\times k}$;
\STATE Initialize $d_{ij}$ as $\|\mathbf{x}_{i}-\mathbf{x}_{j}\|_{2}^{2}$, and $\mathbf{S}$ using (\ref{eq:SDSPCAAN-si});
\STATE $\lambda=1$;
\FOR{$t=1$ to $T$}
\STATE $\mathbf{S}=(\mathbf{S}+\mathbf{S}^T)/2$;
\STATE $\mathbf{L}=\operatorname{diag}(\mathbf{S1}_n)-\mathbf{S}$;
\STATE $\mathbf{Z}=\mathbf{Z}_0+\beta\mathbf{D}+\delta \mathbf{X}\mathbf{X}^T\mathbf{L} \mathbf{X}\mathbf{X}^T$;
\STATE Construct $\mathbf{Q}$ by the $k$ trailing eigenvectors of $\mathbf{Z}$;
\STATE Calculate the $c+1$ smallest eigenvalues of $\mathbf{L}$ in ascending order as $e_1, e_2, \ldots , e_{c+1}$;
\IF{$\sum_{i=1}^c e_i > tol$}
\STATE $\lambda=2\lambda$;
\ELSIF{$\sum_{i=1}^{c+1} e_i < tol$}
\STATE $\lambda=\lambda/2$;
\ELSIF{$\|\mathbf{Q}-\mathbf{Q}_0\|_{1,1} < tol$}
\STATE break;
\ENDIF
\STATE Update $\mathbf{D}$ using (\ref{eq:SDSPCA-V});
\STATE Calculate $d_{ij}=\|\mathbf{Q}^T\mathbf{X}\mathbf{x}_{i}-\mathbf{Q}^T\mathbf{X}\mathbf{x}_{j}\|_{2}^{2}+\lambda\|\mathbf{y}_i-\mathbf{y}_j\|_{2}^{2}$;
\STATE Update $\mathbf{S}$ using (\ref{eq:SDSPCAAN-si});
\STATE $\mathbf{Q}_0=\mathbf{Q}$;
\ENDFOR
\STATE $\mathbf{W}=\mathbf{X}^T\mathbf{Q}$.
\end{algorithmic}
\end{algorithm}

\section{Experiments}

Experiments on nine real-world datasets are performed in this section to demonstrate the performance of the proposed SDSPCAAN.

\subsection{Datasets}

The following nine high-dimensional benchmark classification datasets were used in the experiments:
\begin{enumerate}
\item Musk1 \cite{Dua}, which consists of 476 conformations belonging to 207 musk molecules and 269 non-musk molecules. Each conformation is described by 166 features.
\item MSRA25, which contains 1,799 front-face images of 12 distinct subjects with different background and illumination conditions. In our experiment, all images were resized to 16$\times$16.
\item Palm, which includes 2,000 images of palm prints from 100 distinct individuals. In our experiment, all images were down-sampled to 16$\times$16.
\item USPST, which contains 2,007 images of handwritten digits from 0 to 9. This dataset was sampled from the original USPS dataset. In our experiment, all images were down-sampled to 16$\times$16.
\item Isolet \cite{Dua}, which contains 1,560 samples from 30 subjects who spoke the name of each alphabet letter twice. Each sample is described by 617 features.
\item Yale, which contains 165 gray-scale face images of 15 distinct subjects. Each subject has 11 images with different facial expressions or configurations: center-light, with glasses, happy, left-light, without glasses, normal, right-light, sad, sleepy, surprised, and wink. In our experiment, all images were down-sampled to 32$\times$32.
\item ORL, which contains 400 face images from 40 distinct subjects. Each subject has 10 images with varying shooting time, lighting, facial expressions and facial details. In our experiment, all images were down-sampled to 32$\times$32.
\item COIL20 \cite{Nene1996}, which contains 1,440 gray-scale images from 20 distinct objects. Each object has 72 images taken at pose interval of 5 degrees. In our experiment, all images were down-sampled to 32$\times$32.
\item YaleB, which contains 2,414 near frontal face images from 38 distinct subjects. Each subject has 64 images under different illuminations. In our experiment, all images were cropped and resized to 32$\times$32.
\end{enumerate}

A summary of the nine datasets is shown in Table~\ref{tab:data}.

\begin{table}[htpb] \centering
\caption{Summary of the nine high-dimensional classification datasets.} \label{tab:data}
\begin{threeparttable}
\begin{tabular}{cccc} \toprule
Dataset & No. of Samples & No. of Features & No. of Classes \\ \midrule
Musk1\tnote{1}   & 476            & 166             & 2              \\
MSRA25\tnote{2}  & 1,799           & 256             & 12             \\
Palm\tnote{2}    & 2,000           & 256             & 100            \\
USPST\tnote{2}   & 2,007           & 256             & 10             \\
Isolet\tnote{3}  & 1,560           & 617             & 2              \\
Yale\tnote{4}    & 165            & 1,024            & 15             \\
ORL\tnote{5}     & 400            & 1,024            & 40             \\
COIL20\tnote{6}  & 1,440           & 1,024            & 20             \\
YaleB\tnote{7}   & 2,414           & 1,024            & 38             \\ \bottomrule
\end{tabular}
\begin{tablenotes}
\item[1] http://archive.ics.uci.edu/ml/datasets/musk+(version+1)
\item[2] http://www.escience.cn/people/fpnie/index.html
\item[3] http://archive.ics.uci.edu/ml/datasets/ISOLET
\item[4] http://www.cad.zju.edu.cn/home/dengcai/Data/Yale/Yale\_32x32.mat
\item[5] http://www.cad.zju.edu.cn/home/dengcai/Data/ORL/ORL\_32x32.mat
\item[6] http://www.cad.zju.edu.cn/home/dengcai/Data/COIL20/COIL20.mat
\item[7] http://www.cad.zju.edu.cn/home/dengcai/Data/YaleB/YaleB\_32x32.mat
\end{tablenotes}
\end{threeparttable}
\end{table}

\subsection{Algorithms}

We compared the performance of eight different dimensionality reduction approaches:
\begin{enumerate}
\item \emph{Baseline}, which uses $\mathbf{I}_k$ as the projection matrix, i.e., the first $k$ features are used in classification.
\item \emph{PCA}, the most popular unsupervised dimensionality reduction approach, introduced in Section~\ref{sect:PCA}.
\item \emph{JPCDA} \cite{Zhao2019a}, which unifies PCA and LDA. It first performs PCA to reduce the feature dimensionality to $k$, then LDA to further reduce the feature dimensionality to $c$. The two steps are optimized simultaneously.
\item \emph{SDSPCA} \cite{Feng2019}, a supervised sparse extension of PCA, introduced in Section~\ref{sect:SDSPCA}. It was implemented by setting $\delta$ in SDSPCAAN to zero.
\item \emph{SLNP} \cite{Pang2019}, a supervised version of PCAN. SLNP learns the class-wise similarity graphs and the projection matrix simultaneously.
\item \emph{SPCAN}, a supervised PCAN, implemented by removing the first three terms in the objective function of (\ref{eq:SDSPCAAN}).
\item \emph{SDSPCA-LPP}, a combination of SDSPCA and LPP, implemented by fixing the similarity graph $\mathbf{S}$ in SDSPCAAN at its initial value.
\item \emph{SDSPCAAN}, our proposed algorithm, introduced in Section~\ref{sect:SDSPCAAN}.
\end{enumerate}

A comparison of the eight algorithms is shown in Table~\ref{tab:algs}.

\begin{table}[htpb] \centering \setlength{\tabcolsep}{0.5mm}
\caption{Comparison of the eight algorithms.} \label{tab:algs}
\begin{threeparttable}
\begin{tabular}{c|cccc} \toprule
\multirow{2}{*}{Algorithm} & Preserve Global  & \multicolumn{1}{|c|}{Preserve Local} & \multicolumn{1}{c|}{Adaptive} & \multirow{2}{*}{Supervised}\\
 & Data Structure  & \multicolumn{1}{|c|}{Data Structure} & \multicolumn{1}{c|}{Neighborhood} &\\ \midrule
Baseline   & --           & --            & --     & --         \\
PCA  & $\checkmark$           &    --          &   --   &  --      \\
JPCDA    & $\checkmark$          &    --         &--  & $\checkmark$            \\
SDSPCA   & $\checkmark$          &  --            & --     &$\checkmark$        \\
SLNP  &   --         & $\checkmark$        & $\checkmark$        &$\checkmark$      \\
SPCAN    &   --          & $\checkmark$       &$\checkmark$ & $\checkmark$             \\
SDSPCA-LPP     & $\checkmark$      & $\checkmark$            &-- & $\checkmark$             \\
SDSPCAAN  & $\checkmark$           & $\checkmark$      &$\checkmark$ &$\checkmark$      \\ \bottomrule
\end{tabular}
\end{threeparttable}
\end{table}

\subsection{Experimental Setup}

We used $1$-nearest neighbor based on standardized Euclidean distance (so that PCA and vPCA are equivalent) as the base classifier. The subspace dimensionality $k$ was tuned from $\{10,20,\ldots,100\}$, with the constraint that $k$ must be no larger than $n$ and $d$, and no smaller than $c$. We set $\epsilon=2^{-52}=2.2204\times 10^{-16}$ (\emph{eps} in Matlab), $tol=10^{-3}$, and $T=500$ for all iterative approaches (JPCDA, SDSPCA, SLNP, SPCAN, SDSPCA-LPP and SDSPCAAN). For JPCDA, $\eta$ was tuned from $\{0.01,0.1,1,10,100\}$. For SDSPCA, SDSPCA-LPP and SDSPCAAN, $\alpha$ and $\beta$ were tuned from $\{0.01,0.1,1,10,100\}\cdot \operatorname{Tr}(\mathbf{X}\mathbf{X}^T) / \operatorname{Tr}(\mathbf{Y}\mathbf{Y}^T)$ and $\{0.01,0.1,1,10,100\}\cdot \operatorname{Tr}(\mathbf{X}\mathbf{X}^T) / \operatorname{Tr}(\mathbf{D})$, respectively. For SDSPCA-LPP and SDSPCAAN, $\delta$ was also tuned from $\{0.01,0.1,1,10,100\}\cdot \operatorname{Tr}(\mathbf{X}\mathbf{X}^T) / \operatorname{Tr}(\mathbf{X}\mathbf{X}^T\mathbf{L}\mathbf{X}\mathbf{X}^T)$.

We randomly partitioned each dataset into three subsets: 20\% for training, 40\% for validation, and the remaining 40\% for test. We repeated this process 10 times for each of the nine datasets, and recorded the test balanced classification accuracies (BCAs; the average of the per-class classification accuracies) \cite{Wu2016} as our performance measure.

\subsection{Experimental Results}

The mean and standard deviation of the test BCAs in 10 runs are shown in Table~\ref{tab:avgBCAs}. The largest value (best performance) on each dataset is marked in bold. %We also show the ranks of the average test BCAs on each dataset in Table~\ref{tab:rankBCAs}.
Note that SLNP cannot run on datasets Palm, Yale and ORL, because there are no adequate samples in each class.

\begin{table*}[htpb] \centering
\caption{Mean and standard deviation of BCAs(\%) of the eight approaches on the nine datasets.} \label{tab:avgBCAs}
\begin{tabular}{@{}ccccccccc@{}}
\toprule
Dataset & Baseline                & PCA            & JPCDA                   & SDSPCA         & SLNP                    & SPCAN                   & SDSPCA-LPP                & SDSPCAAN                 \\ \midrule
Musk1   & \textbf{79.47$\pm$2.35} & 76.03$\pm$3.75 & 74.06$\pm$4.14          & 77.32$\pm$1.89 & 66.78$\pm$4.47          & 71.33$\pm$4.49          & 78.04$\pm$2.59          & 77.31$\pm$3.24          \\
MSRA25  & 98.15$\pm$0.61          & 98.89$\pm$0.45 & 99.73$\pm$0.14          & 99.77$\pm$0.20 & 99.76$\pm$0.17          & 43.36$\pm$11.54         & 99.68$\pm$0.32          & \textbf{99.78$\pm$0.13} \\
Palm    & 90.58$\pm$1.65          & 96.71$\pm$0.80 & \textbf{97.36$\pm$0.97} & 96.94$\pm$0.92 & --          & 61.45$\pm$3.15          & 97.14$\pm$1.01          & 96.88$\pm$0.92          \\
USPST   & 68.06$\pm$1.69          & 85.28$\pm$1.45 & 86.04$\pm$1.48          & 87.76$\pm$0.81 & 81.66$\pm$1.71          & 17.91$\pm$2.49          & \textbf{87.89$\pm$1.22} & 87.52$\pm$1.00          \\
Isolet  & 66.01$\pm$1.95          & 83.98$\pm$2.01 & 81.19$\pm$1.56          & 84.26$\pm$1.56 & \textbf{91.78$\pm$1.39} & 75.73$\pm$2.00          & 86.29$\pm$2.01          & 86.66$\pm$2.40          \\
Yale    & 23.57$\pm$6.31          & 40.44$\pm$4.55 & 44.67$\pm$5.87          & 39.83$\pm$3.55 & --          & 47.16$\pm$4.82          & 38.95$\pm$3.74          & \textbf{48.24$\pm$4.40} \\
ORL     & 30.56$\pm$4.00          & 58.32$\pm$5.19 & 61.85$\pm$3.76          & 63.35$\pm$4.21 & --          & \textbf{69.70$\pm$5.71} & 63.72$\pm$4.71          & 69.69$\pm$5.78          \\
COIL20  & 62.29$\pm$1.58          & 93.21$\pm$1.63 & 94.90$\pm$0.98          & 94.42$\pm$1.54 & 92.78$\pm$0.96          & 92.77$\pm$1.51          & 95.80$\pm$0.59          & \textbf{97.12$\pm$0.96} \\
YaleB   & 50.07$\pm$1.70          & 78.94$\pm$0.96 & \textbf{83.71$\pm$1.37} & 78.84$\pm$0.96 & 80.86$\pm$1.06          & 78.93$\pm$1.80          & 79.88$\pm$1.55          & 80.08$\pm$1.13          \\ \midrule
Average & 63.20$\pm$1.02          & 79.09$\pm$0.67 & 80.39$\pm$1.17          & 80.27$\pm$0.63 & --          & 62.04$\pm$1.94          & 80.82$\pm$0.65          & \textbf{82.59$\pm$0.92} \\ \bottomrule
\end{tabular}
\end{table*}

%\begin{table*}[htpb] \centering
%\caption{BCA ranks of the eight approaches on the nine datasets.} \label{tab:rankBCAs}
%\begin{tabular}{@{}ccccccccc@{}}
%\toprule
%Dataset & Baseline & PCA & JPCDA & SDSPCA & SLNP & SPCAN & SDSPCA-LPP & SDSPCAAN \\ \midrule
%Musk1   & 1        & 5   & 6     & 3      & 8    & 7     & 2        & 4       \\
%MSRA25  & 7        & 6   & 4     & 2      & 3    & 8     & 5        & 1       \\
%Palm    & 6        & 5   & 1     & 3      & 8    & 7     & 2        & 4       \\
%USPST   & 7        & 5   & 4     & 2      & 6    & 8     & 1        & 3       \\
%Isolet  & 8        & 5   & 6     & 4      & 1    & 7     & 3        & 2       \\
%Yale    & 7        & 4   & 3     & 5      & 8    & 2     & 6        & 1       \\
%ORL     & 7        & 6   & 5     & 4      & 8    & 1     & 3        & 2       \\
%COIL20  & 8        & 5   & 3     & 4      & 6    & 7     & 2        & 1       \\
%YaleB   & 8        & 5   & 1     & 7      & 2    & 6     & 4        & 3       \\ \midrule
%Average & 6.6      & 5.1 & 3.7   & 3.8    & 5.6  & 5.9   & 3.1      & 2.3     \\ \bottomrule
%\end{tabular}
%\end{table*}

Table~\ref{tab:avgBCAs} shows that:
\begin{enumerate}
\item Our proposed SDSPCAAN performed the best on three out of the nine datasets, and close to the best on the remaining six datasets. On average, SDSPCAAN performed the best.
\item SDSPCAAN outperformed SDSPCA on six out of the nine datasets, and slightly under-performed SDSPCA on the remaining three datasets. These results suggested that the features learnt by SDSPCA may not be adequate since it did not utilize the local data structure information, which is particularly evident on the Yale and ORL datasets.
\item SDSPCAAN outperformed SPCAN on eight of the nine datasets, and slightly under-performed SPCAN on the remaining one dataset. These results suggested that the futures learnt by SPCAN may not be adequate, since it did not utilize the global data structure information, which is particularly evident on the MSRA25 and USPST datasets.
\item SDSPCAN outperformed SDSPCA-LPP on six of the nine datasets, and under-performed it on the remaining three datasets. These indicated that the fixed similarity graph in SDSPCA-LPP may lead to suboptimal results, and our proposed SDSPCAAN can improve it by effectively utilizing local data structure information through adaptive-neighborhood.
\end{enumerate}

In summary, SDSPCAAN outperformed other state-of-the-art dimensionality reduction approaches, because it can effectively utilize both global and local data structure information by combining SDSPCA and PCAN.

\subsection{Effect of the Subspace Dimensionality}

\begin{figure}[!h] \centering
\includegraphics[width=.7\linewidth,clip]{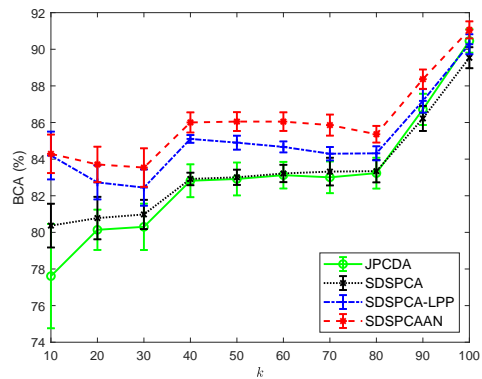}
\caption{BCA versus the subspace dimensionality $k$.} \label{fig:BestBCAVSKerr}
\end{figure}

To study the effect of the subspace dimensionality $k$, we varied $k$ in $[10, 100]$ while keeping other parameters ($\alpha$, $\beta$ and $\delta$) at their best value, and recorded the averaged test BCA of all nine datasets, as shown in Fig.~\ref{fig:BestBCAVSKerr}. For $k\in[10,100]$, our proposed SDSPCAAN always outperformed the state-of-the-art JPCDA and SDSPCA. This again indicated that SDSPCAAN can effectively utilize both global and local data structure information, by combining SDSPCA and PCAN.

\subsection{SDSPCAAN Parameters Sensitivity}

\begin{figure*}[!h] \centering
\subfigure[]{\label{fig:BestBCAVSp1}     \includegraphics[width=.32\linewidth,clip]{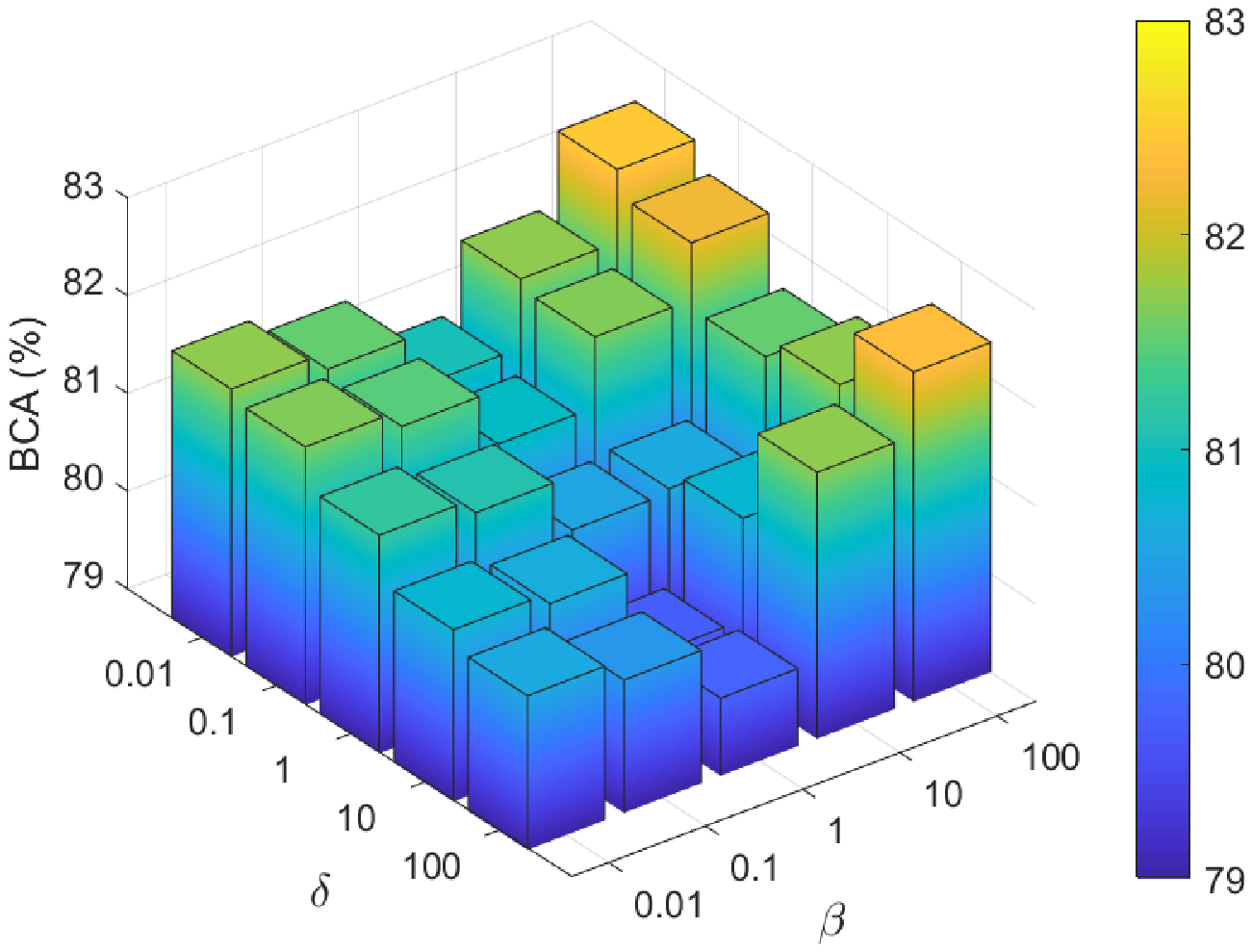}}
\subfigure[]{\label{fig:BestBCAVSp2}     \includegraphics[width=.32\linewidth,clip]{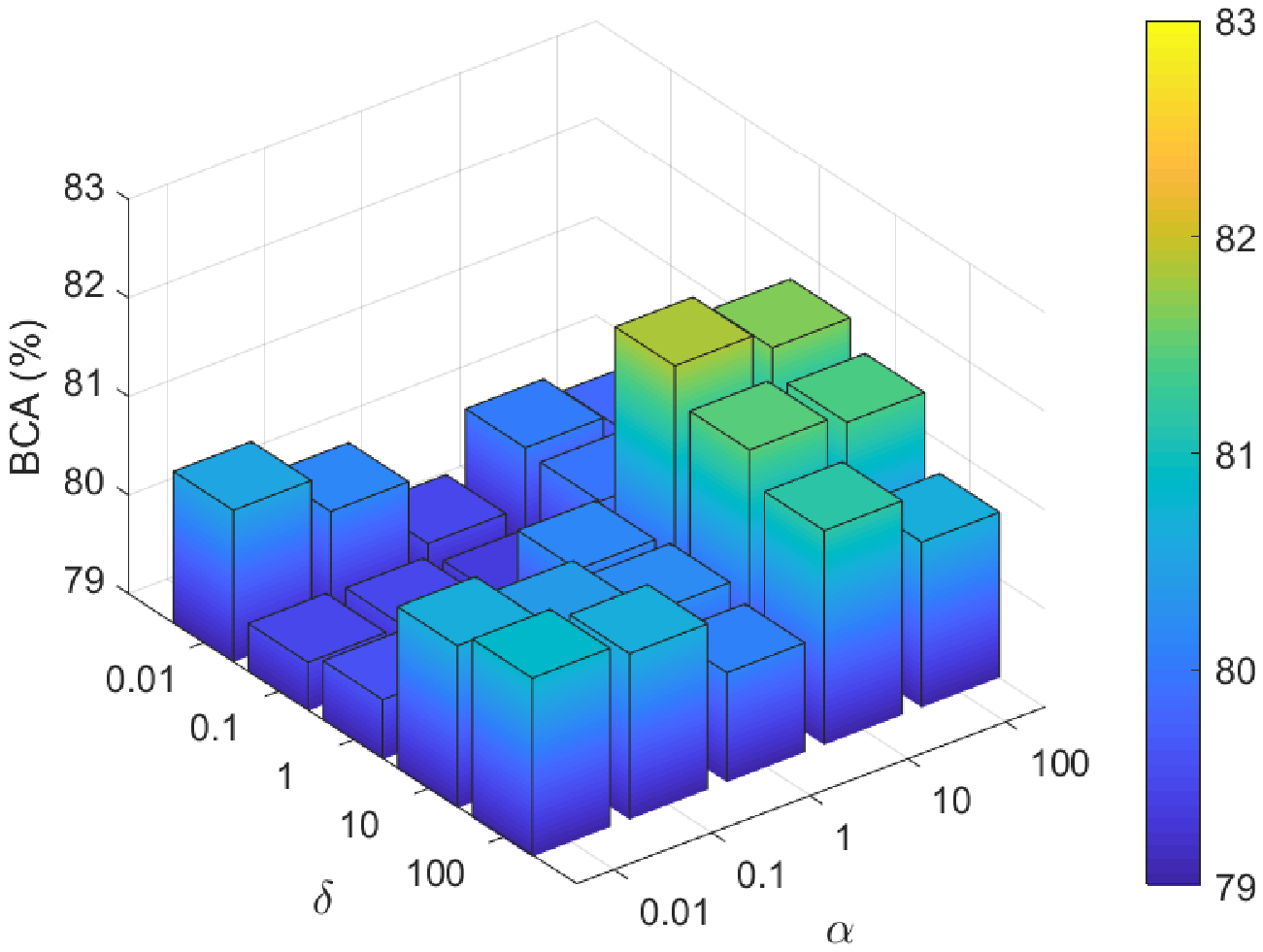}}
\subfigure[]{\label{fig:BestBCAVSp3}     \includegraphics[width=.32\linewidth,clip]{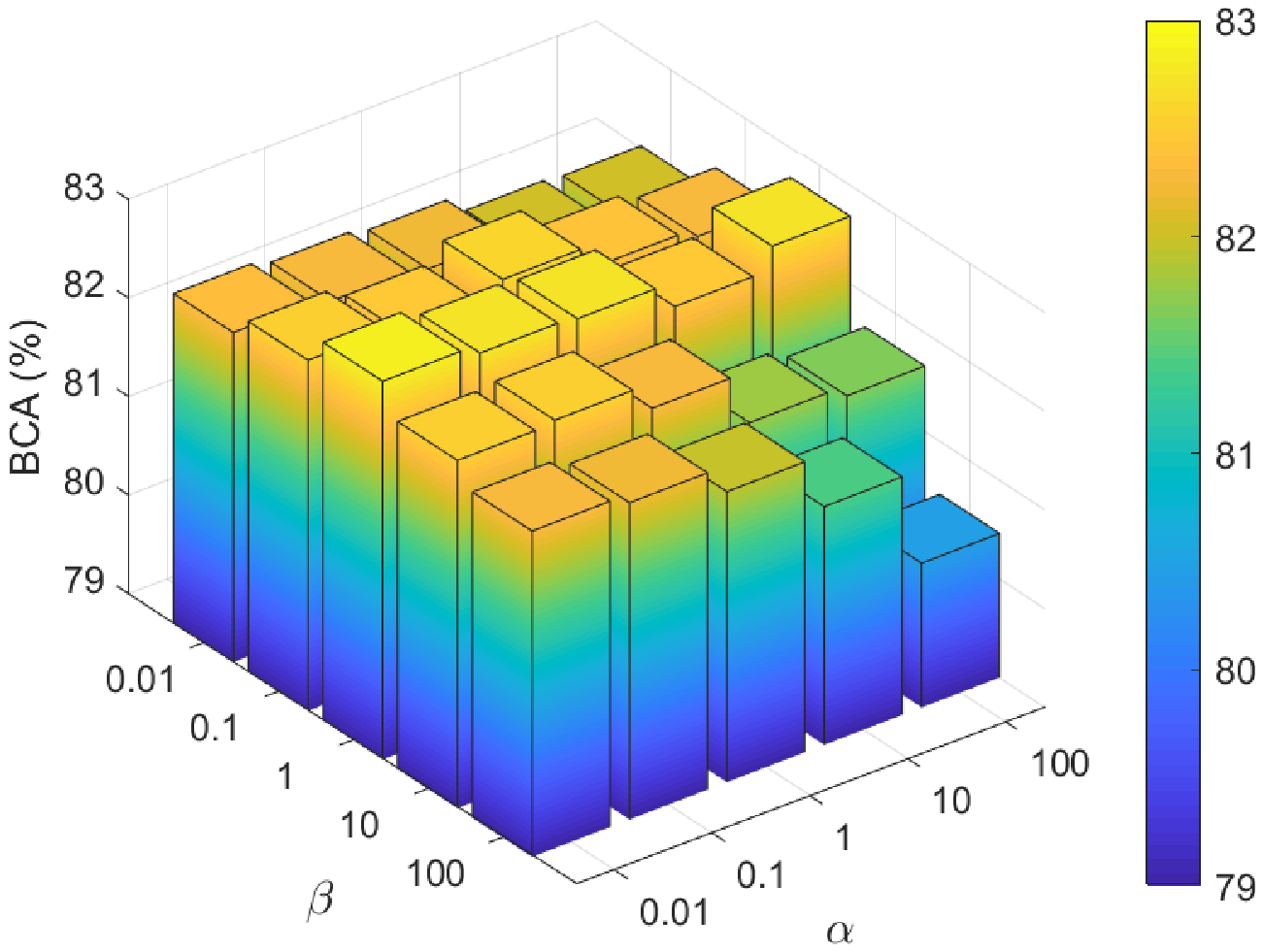}}
\caption{BCA of SDSPCAAN versus its parameters. (a) $\beta$ and $\delta$; (b) $\alpha$ and $\delta$; (c) $\alpha$ and $\beta$.} \label{fig:BestBCAVSp}
\end{figure*}

SDSPCAAN  has three parameters, $\alpha$, $\beta$ and $\delta$. It is important to analyze how these parameters affect its performance. The results are shown in Fig.~\ref{fig:BestBCAVSp}. Take Fig.~\ref{fig:BestBCAVSp1} as an example. We changed $\beta$ and $\delta$, while keeping other parameters ($k$ and $\alpha$) at their best value, and recorded the averaged test BCA of all nine datasets. We may conclude that SDSPCAAN is robust to $\alpha$ and $\beta$ in the range $[0.01,100]$, but sensitive to $\delta$.

\section{Conclusion}

In this paper, we have proposed a novel linear dimensionality reduction approach, SDSPCAAN, that unifies SDSPCA and PCAN to extract the most discriminant features for classification. Our experiments demonstrated that SDSPCAAN can effectively utilize both global and local data structure information in dimensionality reduction, and learning the similarity graph from adaptive neighbors can further improves its performance. When the extracted features were used in a 1-nearest neighbor classifier, SDSPCAAN outperformed several state-of-the-art linear dimensionality reductions approaches.


\end{document}